# Learning to replenish: A hybrid deep reinforcement learning for dynamic inventory management in the pharmaceutical supply chains

**Amandeep Kaur***
Department of Management Studies
ABV-Indian Institute of Information Technology and Management Gwalior-474015, Madhya Pradesh, India
Email: adeep5524@gmail.com
Mobile: +91-9780566060
Orcid Id: 0000-0002-8325-668X
(* Corresponding Author)

**Gyan Prakash**
Department of Management Studies
ABV-Indian Institute of Information Technology and Management Gwalior-474015, Madhya Pradesh, India
Email: gyanprakasha@yahoo.com
Mobile: +91-9406581635

**Statements and Declarations**

**Competing Interests:** The authors declare that they have no known competing financial interests or personal relationships that could have appeared to influence the work reported in this paper.

# Learning to replenish: A hybrid deep reinforcement learning for dynamic inventory management in the pharmaceutical supply chains

***Abstract—*** Pharmaceutical supply chains (PSCs) struggle with inventory management (IM) due to unpredictable demand patterns and variable lead times associated with restocking. This complexity is further compounded by the finite shelf lives of pharmaceutical products, which necessitate a delicate balance between adequate stock and minimal waste. These intertwined factors create a complex optimization problem that requires sophisticated inventory strategies to ensure both product availability and PSC efficiency. This study aims to develop an optimal inventory replenishment policy for pharmaceutical products that can handle the stochasticity arising from uncertain demand and variable PSC conditions. The objective is to maximize the profitability of the PSC while maintaining a high patient service level. We formulate the problem as a Markov decision process and propose a deep reinforcement learning (DRL) approach, specifically, a hybrid asynchronous advantage actor–critic distributed proximal policy optimization (A3C-DPPO) algorithm. The A3C-DPPO algorithm is tailored to handle the continuous action space inherent in IM. The numerical results demonstrate that the proposed algorithm adaptively updates the inventory replenishment strategy under dynamic scenarios, resulting in lower inventory costs compared to various benchmarks. We also conduct numerical validation using real-world pharmaceutical inventory data to confirm the practical feasibility of the proposed algorithm.

## 1. Introduction

Stockouts of life-saving medicines can have serious consequences, including patient deaths. SDG3 of the United Nations' Sustainable Development Goals mandates “access to safe, effective, quality, and affordable essential medicines and vaccines for all” by 2030 (Karamshetty et al., 2022). Medications have limited shelf lives, contain active molecular ingredients that degrade over time, and often require specific storage conditions (Laínez et al., 2012). To deliver a very high patient service level (PSL) of almost 100%, the pharmaceutical supply chain (PSC) maintains high inventory levels (Li et al., 2018) (Priyan and Mala, 2019). Inventory management (IM) along the PSC aims to deliver the right amount of medication at the right place, at the right time, and at an affordable price (Bozkir et al., 2022). IM in PSCs has considerable economic implications: global healthcare market revenues were USD 2.0 billion in 2022 and are estimated to reach USD 5.0 billion by the end of 2035 (Research Nester 2023). Global healthcare expenditure increased from 8.8% of GDP in 2019 (Issever et al., 2024) to 10.8% (USD 9 trillion) in 2020 during the COVID-19 pandemic. The share of these expenditures on pharmaceuticals varied from 5% to 50%, depending on the country's income level (WHO, 2022).

IM is a continuous, sequential process involving interconnected decisions influenced by factors such as demand uncertainty, operational inefficiencies, delivery lead times, and storage requirements (Wong et al., 2023). Traditional inventory replenishment policies often fall short in adapting to the highly complex and dynamic nature of PSCs. Multi-echelon PSCs comprise networks with multiple distribution stages from point-of-origin to patient, and the complexity of these networks is further heightened by fluctuating demand, variable lead times, poor cooperation and information asymmetry (Govindan et al., 2023, Shang et al., 2022). Financial factors, upstream and downstream characteristics, intra-organizational capabilities, inter-organizational cooperation, and the business environment also impact PSCs (Hansen et al., 2023). Demand–supply mismatches and frequent stockouts result in increased morbidity and mortality. Emergency orders not only incur high additional costs but may also result in waste and disposal issues.

Machine learning (ML), particularly reinforcement learning (RL), has emerged as a powerful tool to tackle inventory issues along supply chains (SCs) (Kim et al., 2024; Rolf et al., 2023; Qian et al., 2022). RL models complex inventory decisions primarily as Markov decision processes (MDPs) and optimization strategies through trial and error. Classical RL algorithms such as Q-learning and state–action–reward–(next) state–(next) action (SARSA) have been applied to manage perishable inventory under fluctuating demand and deterministic lead times (Saha et al., 2024; Kara & Dogan 2018). However, these approaches typically operate within discrete state–action spaces, limiting their effectiveness in complex real-world applications. As IM becomes more complex, classical RL algorithms are increasingly susceptible to the curse of dimensionality and struggle with large state–action spaces.

To circumvent the curse of dimensionality and solve complex MDPs, deep reinforcement learning (DRL) employs deep neural networks to approximate the value or policy functions of MDPs (Gijsbrechts et al., 2022). The two primary DRL approaches are value-based and policy-based. The goal of value-based approaches is to find the optimal value function that quantifies the expected cumulative rewards associated with specific actions in given states. Policy-based approaches, such as policy gradient or actor–critic, aim to directly identify an optimal policy that specifies the best action to take in a given state. Policy-based approaches focus on improving the policy itself and are often best-suited for environments with continuous action spaces or high-dimensional state spaces.

Several researchers have developed DRL-based solutions for effective IM in PSCs (Oroojlooyjadid et al. 2022, Moor et al. 2022, Ahmadi et al 2022). For instance, Ahmadi et al. (2022) used Q-learning and deep Q-network (DQN) to develop intelligent IM approaches for managing perishable pharmaceutical products, considering uniform demand patterns in hospitals and central warehouses, and Boute et al. (2022) proposed DRL-based SC optimization for predicting demand patterns and lead times. However, most previous solutions have focused on either discrete or quantized action spaces, reducing their ability to capture highly complex scenarios. Furthermore, the use of a uniform distribution to model explicit demand patterns limits their applicability to dynamic SC contexts. For stochastic demands, continuous action spaces may offer more effective solutions than discrete action spaces (Peng et al., 2019), but the use of continuous action spaces to develop solutions with policy-based DRL algorithms is limited (Geevers et al., 2023; Vanvuchelen et al., 2020), particularly for PSC IM.

This study addresses key challenges in PSCs by proposing a novel hybrid asynchronous advantage actor–critic distributed proximal policy optimization (A3C-DPPO) algorithm for inventory replenishment. The algorithm is designed to operate in a continuous action space, enabling precise control over order quantities and dynamic inventory optimization. Integrating actor–critic methods with the policy optimization mechanisms of PPO enhances exploration efficiency in complex pharmaceutical environments. Our approach emphasizes stakeholder coordination between distribution centers and retailers to optimize IM. The findings provide valuable insights for improving SC efficiency, reducing costs, and maintaining high service levels in pharmaceutical networks.

The remainder of this article is organized as follows: Section 2 presents a comprehensive review of the literature on IM to highlight the research gap. Section 3 formulates the mathematical model within the multi-echelon PSC environment. In Section 4, the hybrid A3C-DPPO algorithm for IM is proposed. Section 5 presents the experimental results, sensitivity analysis, and numerical validation using real-world data, which is followed by a discussion of managerial and sustainability implications. Finally, Section 6 concludes the study and outlines the scope for future research.

## 2. Related work

We first review IM techniques, from classical methods to AI/ML approaches, with a particular focus on their relevance to PSCs. Our aim is to highlight the potential of these approaches to address the unique challenges posed by the pharmaceutical sector, including demand fluctuations, perishability, and multi-echelon distribution.

### *2.1 Classical IM approaches*

Classical IM often relies on deterministic models in which demand, lead time, and other factors are relatively stable and known. Approaches such as fuzzy programming, stochastic programming, dynamic optimization, and probabilistic optimization have been used to optimize inventory levels and reduce costs. To highlight the competitive behaviors between pharmaceutical companies that influence inventory decisions, Chung & Kwon (2016) explored integrated SC for perishable pharmaceutical products considering dynamic and oligopolistic competition. Zahiri et al. (2018) considered demand and cost uncertainties, product shelf life, sustainability, and quantity discounts to formulate a multi-objective optimization problem and proposed a Pareto-optimal solution with a possibilistic optimization approach. Elarbi et al. (2021) developed a stochastic, multi-period, multi-product mathematical model by assuming normally distributed stochastic demand for IM along a multi-echelon PSC in Tunisia. Candan et al. (2016) formulated the inventory problem in the PSC as a mixed integer linear programming (MILP) model and proposed a supplier-driven vendor-managed inventory (VMI) system for planning, scheduling, and cost optimization. Weraikat et al. (2019) extended the VMI system by proposing a mixed-integer non-linear programming (MINLP) model aimed at minimizing the quantity of expired medications and avoiding shortages while maintaining required inventory levels at hospitals under deterministic demand.

PSL is a crucial PSC factor that directly impacts patient care and operational efficiency (Neve & Schmidt, 2022; Bijvank & vis 2012; Saha & Rathore 2024). In PSCs, PSL is often quantified as the percentage of demand met without stockouts over a period. Saha and Ray (2019) proposed a patient condition-based IM system that dynamically adjusts inventory levels based on patients' conditions to ensure the availability of critical medications. Kochakkashani et al. (2023) underscored the need for high PSL during crises such as COVID-19 and developed MILP model for PSCs that considers uncertainties in demand, supply, and transportation as an intrinsic part of the problem and addresses them using the wait-and-see method. Their model clusters drugs and vaccines based on their merits and drawbacks using K-means clustering by assuming deterministic lead times and an average demand scenario. Bozkir et al. (2022) proposed stocking and transshipment policies to manage pharmaceutical inventories during shortages with the aim of optimizing PSL and ensuring that critical medications remain available to patients. However, these state-of-the-approaches, while addressing issues related to planning, inventory costs, and demand in PSCs have certain limitations. Many were not developed for multi-echelon SC environments involving multiple stakeholders, and their implementation in real-time environments is often not feasible.

### *2.2 Artificial intelligence (AI) and ML-based IM approaches*

The integration of AI and ML into IM has led to significant advances in multi-echelon SC management that leverage data-driven decision-making and optimization techniques (Saha & Rathore, 2024; Ahmadi et al. 2022). Two major categories of AI-based inventory models are (i) ML-based models that learn demand patterns and optimize inventory decisions and (ii) metaheuristic optimization-based models that efficiently search for optimal inventory policies under complex constraints.

Masoudi & Mirzazadeh (2022) proposed an AI-based inventory model incorporating metaheuristic optimization techniques such as particle swarm optimization (PSO) and genetic algorithms (GAs). These methods optimize inventory policies by navigating large solution spaces efficiently, addressing uncertainties in deterioration rates and lead times. Similarly, Nasrollahi & Razmi (2021) developed a multi-objective optimization model to manage demand uncertainties and minimize total costs in a four-layer multi-echelon PSC. Their approach leveraged a non-dominated ranked GA to generate Pareto-optimal solutions. Despite their advantages, metaheuristic techniques often struggle when applied to large and complex SC networks. Additionally, tuning these algorithms to specific SC conditions is challenging. To address these limitations, Lofti et al. (2022) explored VMI strategies for inventory cost reduction and proposed a robust optimization model that integrates data-driven and fuzzy approaches to effectively manage demand uncertainty.

In complex multi-echelon SCs, RL excels at solving sequential decision-making problems where complete knowledge of the underlying stochastic processes is often unavailable. Instead of relying on predefined models, RL agents interact with the SC environment to gather information and optimize decisions over time (Wu et al., 2023; Preil & Krapp, 2022). For instance, Rana & Oliveira (2015) used Q-learning with eligibility traces to model optimal pricing of interdependent perishable products under a stochastic demand pattern. Kara and Dogan (2018) considered the high demand variance and short shelf life of perishable products as important parameters and proposed two different ordering policies: stock-based and age-based. They employed value-based Q-learning and SARSA to address ordering decisions in a single-product single-echelon inventory system, but the algorithm has poorer performance in large state–action spaces. To handle non-stationary demands, Park et al. (2023) developed a stochastic replenishment policy with a structured RL algorithm for single-item and multi-item systems for seasonal IM. Non-stationary demand followed a regime-switching Gamma distribution, whereas stationary demand followed single values for a Gamma distribution and truncated Normal distribution.

Compared to RL, DRL has greater potential for tackling complex inventory problems characterized by large state–action spaces (Stranieri et al., 2024, Shukla et al. 2025). A thorough overview of the field was previously provided by Boute et al. (2022), who extensively reviewed DRL applications in inventory optimization, logistics, and SC. Ahmadi et al. (2022) managed the inventory of perishable pharmaceutical products with Q-learning and DRL policies to account for demand uncertainties and product shelf life. These value-based RL techniques tend to maximize revenue only under a specific setting of discrete state–action space. Sultana et al. (2020) proposed a multi-agent RL framework to capture the stochastic demand pattern of perishable products for multi-product IM but considered the quantized action space, limiting its performance under complex scenarios. According to Peng et al. (2019), the continuous action space provides more effective solutions under stochastic demand patterns than its discrete counterpart. To address lead time constraints, Meisheri et al. (2022) proposed a scalable approach to multi-product IM with continuous action spaces for replenishment decisions. They offered solutions using DQN and PPO for two distinct datasets of perishable goods in scenarios with no lead time delay or deterministic lead times, but they did not consider scenarios with uncertain demand or stochastic replenishment lead times. Gijsbrechts et al. (2022) implemented the A3C algorithm, a hybrid of policy- and value-based DRL, to optimize inventory decisions across three classic problems: lost sales, dual sourcing, and multi-echelon IM. To address demand fulfillment in online retail while considering stochastic demand and replenishment lead time, Wang and Minner (2024) formulated a semi-MDP to optimize demand fulfillment decisions and developed a single-threaded PPO algorithm to solve the problem.

Table 1 presents a summary of the extant literature on SC and DRL in IM, including studies considering pharmaceutical products. Many of these studies adopt deterministic models and thus neglect the complexities of stochastic demand in many real-world scenarios (Meisheri et al., 2022; Masoudi and Mirzazadeh 2022). Very few address PSL, a critical factor in PSCs, or consider emergency orders, which are crucial for maintaining high PSL (Uthayakumar & Priyan, 2013, Bozkir et al. 2022, Baloch and Gzara 2020). Moreover, most research assumes that demand uncertainty follows well-established probability distributions, such as Normal or Uniform (Ahmadi et al. 2022, Elarbi et al. 2021), which simplifies modeling but may not capture the real-world complexities of unpredictable demand variations, such as sudden surges due to emergencies. Some studies exploring the use of Q-learning and DRL for IM along the SC choose discrete action spaces for replenishment decisions, which limits their flexibility, as replenishment decisions often involve continuous action spaces. Expanding DRL approaches to accommodate continuous action spaces could lead to more practical and robust solutions for PSC IM.

## 3. Problem formulation

Here, we formulate the IM problem within a multi-echelon PSC network that includes pharmaceutical manufacturers, distribution centers, and retailers as key entities. The manufacturing unit is responsible for producing pharmaceutical products, which are shipped through distribution centers to retailers such as pharmacies, hospitals, and physician offices, as depicted in Fig.1. The distribution centers are the sole decision-makers for procurement from the manufacturer and act as suppliers for retailers. The inventory system involves sequential processes of order placement, product delivery, sales, and disposal of expired products.

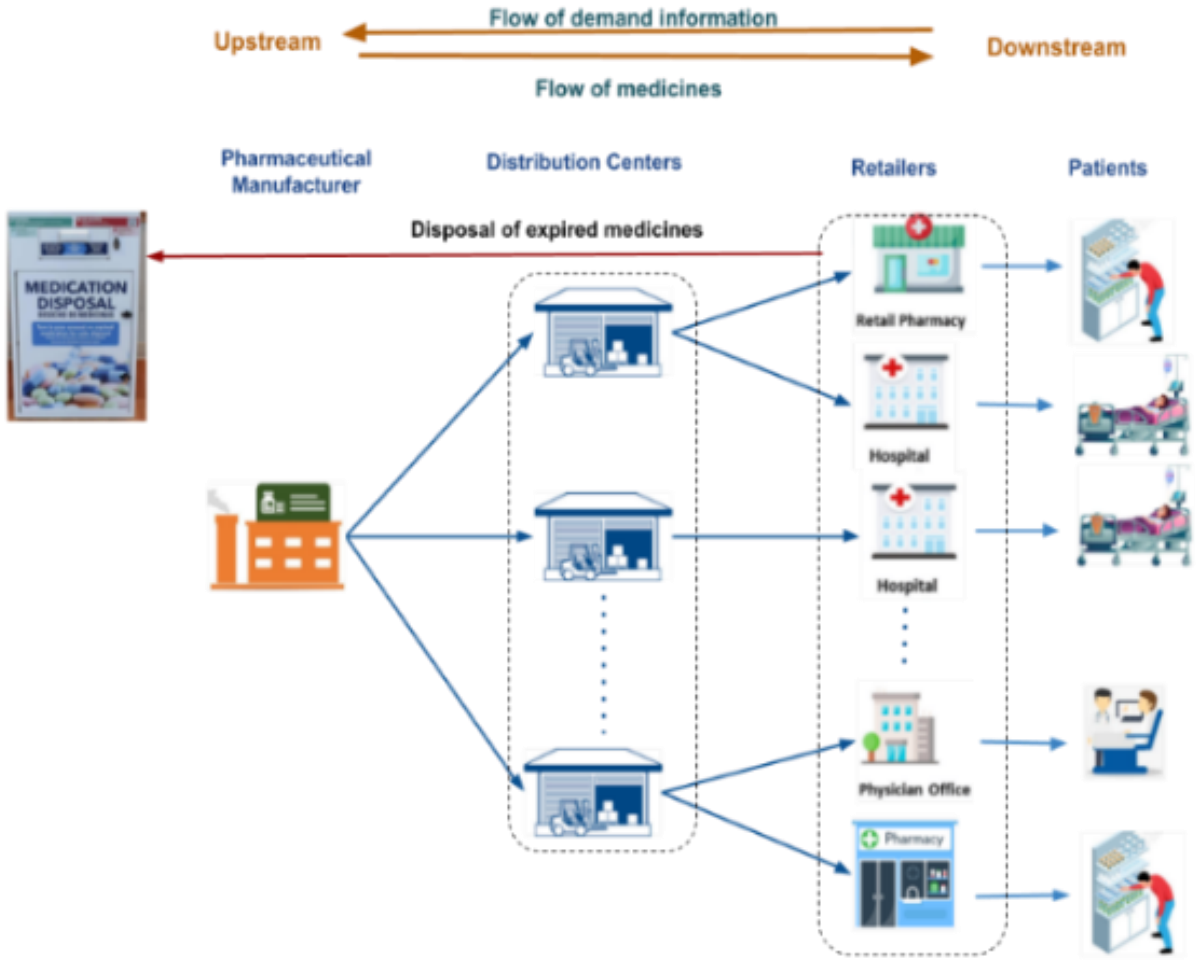

**Fig. 1** A typical multi-echelon and multi-network structure in a PSC

**Table 1**

Summary of the relevant literature

| **Authors** | **Product type** | **Objective function** | **Inventory control policy** | **Modeling approach** | **Solution approach** | **Presence of pharmaceutical product(s)** | **Demand uncertainty modelled as** | **Presence of variables** | | | **Consideration of continuous action space** | **Presence of real-time decision-making** |
|---|---|---|---|---|---|---|---|---|---|---|---|---|
| | | | | | | | | **PSL** | **Emergency orders** | **Replenishment rate** | | |
| Uthayakumar & Priyan (2013) | Multiple | Cost | *(r, Q)* policy | OR-based | Lagrangian multiplier algorithmic approach | √ | Normal distribution | √ | - | √ | - | - |
| Chung & Kwon (2016) | Multiple | Cost | Decentralized replenishment | Game Theory | Nash equilibrium | √ | - | - | - | √ | - | - |
| Candan et al. (2016) | Multiple | Cost | VMI | MILP | Exact | √ | - | - | - | √ | - | - |
| Nematollahi et al. (2017) | Single | Cost | *(T, R)* policy | Mathematical | Approximate | √ | Normal distribution | √ | √ | √ | - | - |
| Zahiri et al. (2018) | Multiple | Cost and Sustainability | *(s, S)* policy | Possibilistic Optimization | Pareto-optimal solutions | √ | Scenario-based | √ | - | √ | - | - |
| Weraikat et al. (2019) | Multiple | Minimize expired stock | VMI | MINLP | Exact | √ | Gamma distribution | √ | - | √ | - | - |
| Baloch & Gzara (2020) | Multiple | Capacity and PSL | Assortment optimization with supplier-driven substitution | MIP | Heuristic | √ | Empirical distributions | √ | - | √ | - | - |
| Priyan & Mala (2020) | Single | Cost | Game-theoretic decision making | CNP | Game theory payoff matrix approach | √ | Normal distribution | √ | - | - | - | - |
| Elarbi et al. (2021) | Multiple | Cost | *(r, Q)* Policy | SP | Mathematical optimization + stochastic simulation | √ | Normal distribution | - | - | √ | - | - |
| Masoudi & Mirzazadeh (2022) | Single | Cost | Heuristic-based demand-driven replenishment | NLP | Metaheuristics | √ | Log-normal | - | - | √ | - | - |
| Nasrollahi & Razmi (2021) | Multiple | Cost & PSL | Demand-driven | Multi-Objective | Exact | √ | Centroid | √ | - | √ | - | - |
| Bozkir et al. (2022) | Multiple | PSL | Inventory pooling | NLP | Simulation | √ | Poisson and exponential | √ | - | - | - | - |
| Ahmadi et al. (2022) | Multiple | Cost | *(R, s, S)* policy | SMIP | QL and DRL | √ | Uniform distribution | √ | √ | √ | - | - |
| Kara & Dogan | Single | Cost | Age-based policy | MDP | QL and SARSA | - | Gamma distribution | - | - | √ | - | - |
| Kochakkashani et al. (2023) | Multiple | Cost and lead time (emergency response) | Cluster-based demand forecasting | MILP + Clustering | Stochastic programming (wait-and-see method) | √ | - | √ | √ | √ | - | - |
| Saha & Rathore (2024) | Multiple | Cost | Non-stationary *(s,c,S',S)* | Semi-MDP | MAB and RL | √ | Negative binomial | √ | √ | √ | - | √ |
| Gijsbrechts et al. (2022) | Multiple | Cost | Dual-sourcing multi-echelon inventory control under lost sales constraints | MDP | DRL | - | Gamma, normal, uniform and geometric distribution | √ | - | √ | - | √ |
| Wang & Minner (2024) | Multiple | Cost | RL-based optimization | Semi-MDP | DRL (PPO+SAC) | - | Compound Poisson demand | √ | √ | √ | √ | √ |
| Geevers et al. (2023) | Multiple | Cost | Base stock policy | MDP | DRL (PPO) | - | Uniform and Poisson distribution | √ | - | √ | √ | √ |
| Sultana et al. (2020) | Multiple | Cost, stock out and waste minimization | RL-based optimization | MDP | DRL (A2C) | - | Binomial distribution | - | - | √ | - | √ |
| **This paper** | Multiple | Cost and PSL | RL-based optimization | MDP | DRL (Hybrid A3C-DPPO) | | Normal, gamma, Poisson and seasonal distribution | √ | √ | √ | √ | √ |

Note: OR: operations research, CNP: constrained non-linear programming, VMI: vendor-managed inventory, MAB: multi-arm bandit, MILP: mixed-integer linear programming, MINLP: mixed integer non-linear programming, NLP: non-linear programming, SMIP: stochastic mixed integer programming, ARM: MDP: Markov decision process, SP: stochastic programming, SARSA: state–action–reward–(next) state–(next) action, PPO: proximal policy optimization, SAC: soft actor–critic, QL: Q-learning, DRL: deep reinforcement learning, A3C-DPPO: asynchronous advantage actor–critic-distributed proximal policy optimization

Initially, each distribution center replenishes its inventory from the pharmaceutical manufacturer, which is assumed to have sufficient capacity to supply as and when needed. The retailers respond to fluctuating medication demands by sourcing from distribution centers. In this PSC, optimal replenishment policies for distribution centers and their associated retailers are necessary.

## *Problem formulation*

Initially, a distribution center, acting as the decision-maker, observes the current inventory level, which includes the on-hand inventory level of each medication $i$ at both the distribution center and the retailers at the beginning of time slot $t$. Let the vectors $\mathbf{x}_d = \left(x_d^1, x_d^2, \ldots\ldots, x_d^M\right)$ and $\mathbf{x}_k = \left(x_k^1, x_k^2, \ldots\ldots, x_k^M\right)$ represent the inventory levels at the distribution center and retailers, respectively, for each medication $i$ at time $t$. The on-hand inventory of medication $i$ at the distribution center and retailers, considering different shelf-life stages, is represented as $x_{d,t}^i = \left(x^{i,1}, x^{i,2}, \ldots\ldots, x^{i,\ell_i}\right)^T$ and $x_{k,t}^i = \left(x^{i,1}, x^{i,2}, \ldots\ldots, x^{i,\ell_i'}\right)^T$, respectively. Here, $\ell_i$ and $\ell'_i$ represent the shelf life of medication $i$ at the distribution center and retailers, respectively, where $\ell'_i < \ell_i$.

**Table 2**
Main notations

| Sets | Definition |
|---|---|
| $t$ | Index of planning periods, $t = 1, 2, \ldots.., T$ |
| $n$ | Index of distribution center/ networks, $n = 1, 2, \ldots.., N$ |
| $k$ | Index of retailers $k = 1, 2, \ldots.., K$ |
| $i$ | Index of medication $i = 1, 2, \ldots\ldots, M$ |
| **Parameters** | |
| $c_{O,d,t}^i$ | Ordering cost per unit of medication $i$ incurred by distribution center at period $t$ |
| $c_{O,k,t}^i$ | Ordering cost per unit of medication $i$ incurred by retailer $k$ at period $t$ |
| $c_{p,k,t}^i$ | Unit shortage penalty cost for medication $i$ at retailer $k$ at period $t$ |
| $c_{H,d,t}^i$ | Inventory holding cost per unit of medication $i$ incurred by distribution center at period $t$ |
| $c_{H,k,t}^i$ | Inventory holding cost per unit of medication $i$ incurred by retailer $k$ at period $t$ |
| $e_{k,t}^i$ | Emergency cost per unit of medication $i$ incurred by retailer $k$ at period $t$ |
| $c_{\varsigma,k}^i$ | Disposal cost for medication $i$ incurred by retailer $k$ |
| $h_{d,t}^i$ | Sales price of medication $i$ at distribution center at period $t$ |
| $h_{k,t}^i$ | Sales price of medication $i$ at retailer $k$ at period $t$ |
| $I_{k,i}^{\max}$ | Maximum inventory storage capacity for medication $i$ at retailer $k$ (in units) |
| $I_{d,i}^{\max}$ | Maximum inventory storage capacity for medication $i$ at distribution center $i$ (in units) |
| $w_k^i$ | Penalty cost per unit of medication $i$ incurred by retailer when $\left(\beta_{k,t}^i < \beta_{k,\min}^i\right)$ |
| $w_d^i$ | Penalty cost per unit of medication $i$ incurred by distribution center when $\left(\beta_{d,t}^i < \beta_{d,\min}^i\right)$ |
| $\beta_{k,\min}^i$ | Minimum acceptable fill rate of medication $i$ at retailer $k$ |
| $\beta_{d,\min}^i$ | Minimum acceptable fill rate of medication $i$ at distribution center |
| $\tau_d^i$ | Expected time of being with current inventory status $x_d^i(t)$ |
| $\tau_k^i$ | Expected time of being with current inventory status $x_k^i(t)$ |

| Decision variables | |
|---|---|
| $x^i_{d,t}$ | On-hand inventory of medication $i$ at distribution center at period $t$ (units) |
| $x^i_{k,t}$ | On-hand inventory of medication $i$ at retailer $k$ at period $t$ (units) |
| $a^i_{d,t}$ | Quantity of medication $i$ ordered by the distribution center from the supplier at period $t$ (units) |
| $a^i_{k,t}$ | Quantity of medication $i$ ordered by retailer $k$ from the distribution center at period $t$ (units) |
| $\ell_i$ | Shelf life of medication $i$ at the distribution center at period $t$ |
| $\ell'_i$ | Shelf life of medication $i$ at retailer $k$ at period $t$ |
| $L^i_{k.t}$ | Replenishment lead time of medication $i$ at the retailer at period $t$ |
| $L^i_{d.t}$ | Replenishment lead time of medication $i$ at the distribution center at period $t$ |
| $D^i_d(t)$ | Demand for medication $i$ at the distribution center at period $t$ |
| $D^i_k(t)$ | Demand for medication $i$ at retailer $k$ at period $t$ |
| $P^i_{k,t}$ | Quantity of medication $i$ expired at retailer $k$ at period $t$ |
| $\delta$ | Binary variable: 1 if an order is placed, and 0 otherwise |

Let the vectors $\mathbf{a}_d = \left(a^1_d, a^2_d, \ldots\ldots, a^M_d\right)$ and $\mathbf{a}_k = \left(a^1_k, a^2_k, \ldots\ldots, a^M_d\right)$ denote the replenishment quantities of medication $i$ arriving at the distribution center and retailers, respectively, at time $t$. Since the distribution center receives supplies from manufacturers under prescheduled contracts with consistent delivery times, its replenishment lead time is assumed to be fixed. By contrast, retailers experience stochastic replenishment lead times due to demand fluctuations and transportation delays, which is best modeled using an exponential distribution as $L^i_k \sim Exp\left(\lambda^i_k\right)$ (Wang & Minner 2024). It is further assumed that all products delivered to the distribution centers have sufficient shelf life, are utilized before expiration, and therefore, incur no disposal costs. During emergencies or stockouts, retailers can place emergency orders to the distribution centers, which are fulfilled immediately. To ensure the timely availability of medications, backorders are not allowed. Key notations used in the problem formulation are summarized in Table 2.

Our objective is to minimize the total incurred costs, including ordering costs, holding costs, stockout penalties, and disposal costs, under stochastic demand while ensuring the required PSL. For retailers, maintaining a high PSL to ensure product availability often leads to increased inventory holding costs and the risk of excess stock, which may result in higher disposal costs for expired medications. Under stochastic demand scenarios, emergency replenishments often incur additional costs due to the urgency of fulfilling the orders. These additional costs serve as a penalty for inadequate inventory planning and can significantly impact profit margins across the SC. The mathematical model incorporating various costs is formulated as follows:

*Distribution center's ordering cost:* $C_1(t) = \sum_{i=1}^{M} c^i_{O,d,t}\,\delta\left(a^i_{d,t}\right)$ (1.1)

*Distribution center's holding cost:* $C_2(t) = \sum_{i=1}^{M} c^i_{H,d,t}\,\tau^i_d\left(a^i_{d,t}\right) \sum_{D^i_{d,t}=0}^{D^i_{d,t}=x^i_{d,t}+a^i_{d,t}} \left(x^i_{d,t} + a^i_{d,t} - D^i_{d,t}\right) p\left(D^i_{d,t}\right)$ (1.2)

Retailer's ordering cost: $C_3(k,t)=\sum_{i=1}^{M} c_{O,k,t}^{i}\delta\left(a_{k,t}^{i}\right)$ (1.3)

Retailer's holding cost: $C_4(k,t)=\sum_{i=1}^{M} c_{H,k,t}^{i}\tau_k^{i}\left(x_{k,t}^{i}\right)\sum_{D_{k,t}^{i}=0}^{x_{k,t}^{i}+a_{k,t}^{i}-D_{k,t}^{i}} P\left(L_k^{i}\le t\right)\left(x_{k,t}^{i}+a_{k,t}^{i}-D_{k,t}^{i}\right)p\left(D_{k,t}^{i}\right)$ (1.4)

Retailer's emergency replenishment cost: $C_5(k,t)=\sum_{i=1}^{M} c_{p,k,t}^{i}\sum_{D_{k,t}^{i}=x_{k,t}^{i}+a_{k,t}^{i}}^{D_{\max,k,t}^{i}}\left(D_{k,t}^{i}-\left(x_{k,t}^{i}+a_{k,t}^{i}\right)\right)p\left(D_{k,t}^{i}\right)p\left(L_{k,t}^{i}>t\right)$ (1.5)

Retailer's disposal cost: $C_6(k,t)=\sum_{i=1}^{M} c_{\varsigma,k}^{i}p_{k,t}^{i}$ (1.6)

Equation (1.1) to (1.6) represents the ordering cost and holding costs, both of which are incurred at the distribution center and retailer levels, along with disposal and emergency replenishment costs at the retailer level. The disposal cost covers the fees for safely discarding expired pharmaceutical products at government-approved sites, while the emergency replenishment cost accounts for the expense of urgent refills for critical medications from the distribution center due to shortages at the retailer level. Equation (1.1) represents the fixed ordering cost specific to each medication, where the binary variable $\delta(.)$ takes a value of 1 if an order is placed and 0 otherwise. Equation (1.2) represents the expected holding cost incurred by the distribution center, which is determined by the duration of time a medication is expected to be in the inventory and the on-hand inventory left after fulfilling the estimated demand for medication $i$ generated from all retailers. It is important to note that the distribution center, $D_{d,t}^{i}$ represents aggregated demand for medication $i$ at time $t$ from all $k$ retailers. The total aggregated demand at distribution center is calculated as $D_{d,t}^{i}=\sum_{k=1}^{K} D_{k,t}^{i}p\left(D_{k,t}^{i}\right)$. Equation (1.3) represents the ordering cost incurred by retailer $k$ when placing an order for medication $i$ at time $t$. Equation (1.4) represents the holding cost incurred by retailer $k$, which depends on the expected duration that medication $i$ remains in inventory. Since the replenishment lead time is variable, the holding cost is influenced by the expected inventory level at time $t$, considering the probability that the replenishment lead time is less than or equal to $t$, indicating that the stock has arrived. Equation (1.5) represents the emergency replenishment cost incurred by retailer $k$ when the on-hand inventory is insufficient to meet the medication demand at a given time $t$, requiring an urgent order from the distribution center. The probability term accounts for the likelihood that the replenishment lead time exceeds $t$, indicating the order has not yet arrived. Equation (1.6) represents the expiration cost incurred by retailer $k$ when a medication reaches the end of its shelf life. At each time step, the shelf life of all medications decreases by 1, and any medications with a shelf life of zero require disposal. The quantity of expired medication at retailer $k$ at the end of the time slot $t$ can be calculated as $p_{k,t}^{i}=\sum_{n=1}^{M_{k,t}^{i}}\mathbf{1}\left(\ell'(t)=0\right)$. Here, $M_{k,t}^{i}$ represents the total number of units of medication $i$ available at retailer $k$ at time $t$ , and $\mathbf{1}(.)$ is an indicator function that equals 1 when the shelf life of the medication reaches zero, indicating that the medication has expired. Finally, the total inventory cost for an SC network consisting of a single distribution center and $k$ retailers in the current period $t$ is computed as the sum of ordering, holding, emergency replenishment, and disposal costs incurred across all entities in the network as

$$C(t)=\sum_{i=1}^{M}\left(C_1(t)+C_2(t)+\sum_{k=1}^{K}\left(C_3(k,t)+C_4(k,t)+C_5(k,t)+C_6(k,t)\right)\right) \tag{2}$$

The objective is to maximize the overall revenue of the PSC while ensuring efficient IM and meeting PSL constraints. Mathematically, for $N$ distribution centers, it can be formulated as,

$$\max\sum_{N}\left\{\sum_{i=1}^{M}\left[\min\left(x_{d,t}^{i}-\sum_{k=1}^{K}a_{k,t}^{i},D_{d,t}^{i}\right)h_{d,t}^{i}+\sum_{k=1}^{K}\min\left(x_{k,t}^{i}+a_{k,t}^{i}+e_{k,t}^{i},D_{k,t}^{i}\right)h_{k,t}^{i}-C_{d,t}^{F,i}-\sum_{k=1}^{K}C_{k,t}^{F,i}\right]-C(t)\right\} \tag{3}$$

where $h_{d,t}^{i}$ and $h_{k,t}^{i}$ denote the sales prices of medication $i$ at the distribution center and at the retailers, respectively.

*Constraints:*

$$x_{d,t+1}^{i}=\max\left(x_{d,t}^{i}+a_{d,t}^{i}-D_{d,t}^{i},0\right)\forall i,t \tag{4}$$

$$x_{k,t+1}^{i}=\max\left(x_{k,t}^{i}+a_{k,t-L_k^i}^{i}+e_{k,t}^{i}-D_{k,t}^{i}-p_{k,t}^{i},0\right)\forall i,k,t \tag{5}$$

$$x_{k,t}^{i}+a_{k,t}^{i}\le I_{k,i}^{\max},\forall i,k,t \tag{6}$$

$$x_{d,t}^{i}+a_{d,t}^{i}\le I_{d,i}^{\max},\forall i,t \tag{7}$$

$$\mathbb{E}\left[\frac{\min\left(x_{k,t}^{i}+a_{k,t}^{i}+e_{k,t}^{i},D_{k,t}^{i}\right)}{D_{k,t}^{i}}\right]\ge\beta_{k,\min}^{i} \tag{8}$$

$$\mathbb{E}\left[\frac{\min\left(x_{d,t}^{i}+a_{d,t}^{i},\sum_{k=1}^{K}D_{k,t}^{i}\right)}{\sum_{k=1}^{K}D_{k,t}^{i}}\right]\ge\beta_{d,\min}^{i} \tag{9}$$

Equation (3) represents the net profit of the PSC, accounting for cash flow, PSL violation penalties, and inventory costs incurred at both the distribution center and retailer levels. Constraints (4) and (5) define the inventory-balancing constraints for the distribution center and retailers. The inventory-balancing constraint indicates that the inventory level at $(t+1)$ depends on the inventory at time $t$ , order replenishment at $t$ , and demand met from the on-hand stock at $t$ . Constraints (6) and (7) represent capacity constraints, which ensure that the inventory levels at the distribution center and retailers do not exceed their respective maximum storage capacities for each medication. Constraints (8) and (9) define the PSL constraints for medications at the distribution center and retailers, respectively. These constraints ensure that both the distribution center and retailers maintain a minimum required Type II service level, which is defined as the expected proportion of demand fulfilled directly from available inventory. If these constraints are not met, a penalty cost is imposed at the distribution center and retailer levels, calculated as $C^{F,i}_{d,t} = w^{i}_{d}.I\left(\beta^{i}_{d,t} < \beta^{i}_{d,min}\right)$ and $C^{F,i}_{k,t}(t) = w^{i}_{k}.I\left(\beta^{i}_{k,t} < \beta^{i}_{k,min}\right)$, where $w^{i}_{d}$ and $w^{i}_{k}$ represent the penalty cost imposed at the distribution center and retailer levels, respectively, based on the risk associated with the unavailability of medication $i$ for patients. The indicator function $I(\beta < \beta_{min})$ equals 1 if the actual fill rate falls below the minimum acceptable fill rate and 0 otherwise.

## 4. Proposed IM framework

### *4.1 RL for IM*

Traditional optimization techniques often fail to adapt dynamically to such uncertainties, making DRL a powerful alternative. To apply DRL algorithms, we formulate the IM problem as an MDP that effectively capture the hierarchical nature of the SC by defining two types of learning agents: a global agent representing the distribution center and multiple local agents representing retailers. The global agent is responsible for managing inventory at the distribution center, placing orders for replenishment from suppliers, and distributing stock among retailers. It optimizes inventory decisions by balancing holding costs, replenishment costs, and PSL constraints while ensuring an adequate supply to meet retailer demand. The local agents (retailers) independently manage their own inventory by deciding how much stock to order from the distribution center and whether to place emergency replenishment orders in case of stock shortages. The goal of the local agents is to minimize stockouts while controlling inventory costs, ensuring that customer demand is met efficiently. To formally address and solve the problem within the context of an MDP, the state space, action space and reward function are defined as follows,

- ***State space***

The state space captures all relevant information from the environment in which the learning agent interacts. Each retailer, acting as an independent agent, interacts with its local environment to determine inventory levels of medication $i$ at retailer $k$; open replenishment orders for medication $i$ placed to the distribution center; emergency replenishment orders for medication $i$; stochastic demand for medication $i$ at retailer $k$; and lead times for replenishment from the distribution center. The state representation of retailer $k$ at time $t$ is

$$u_k(t) = \left\{x^{i}_{k,t}, a^{i}_{k,t}, e^{i}_{k,t}, D^{i}_{k,t}, L^{i}_{k,t}\right\} \tag{10}$$

Similarly, at the global level, the distribution center observes its own environment, collecting information as its state observation. Mathematically,

$$u_d(t) = \left\{x^{i}_{d,t}, a^{i}_{d,t}, \sum_{k=1}^{K} D^{i}_{k,t}, L^{i}_{d,t}\right\} \tag{11}$$

• ***Action space***

Each agent takes actions based on its state observations to optimize IM. The global agent is responsible for managing inventory distribution and supplier replenishment. Based on its state observation, the action taken by global agent is defined as

$$v_d(t) = \left\{a_{d,t}^i, \left\{a_{k,t}^i\right\}_{k=1}^K\right\} \tag{12}$$

Similarly, the action taken by local agent at time $t$, denoted as $v_k(t)$, includes decisions about replenishment orders from distribution centers and the quantity of emergency replenishment. Formally, the action taken by the local agent is defined as

$$v_k(t) = \left\{a_{k,t}^i, e_{k,t}^i\right\} \tag{13}$$

The action space is modeled as a continuous action space, which allows both the distribution center and retailers to make fine-tuned decisions regarding inventory replenishment and allocation. Unlike discrete action spaces, where decisions are limited to predefined levels, continuous action space provides flexibility and more precise inventory control (Vanvuchelen and Boute, 2022; Stranieri and Stella, 2022). To ensure that the agents' actions remain within a feasible action space and prevent negative inventory decisions, we set the lower bound to zero. The upper bound is constrained by the maximum storage capacity, ensuring that inventory levels do not exceed storage limits (Stranieri et al., 2024).

- ***Reward:***

The reward function ensures that agents aim to maximize profit while minimizing inventory replenishment costs and penalty costs associated with stockouts. For a global agent, the reward function is defined as

$$\mathrm{r_d}(\mathrm{t}) = \sum_{\mathrm{i=1}}^{\mathrm{M}}\left[\min\left(\mathrm{x_{d,t}^i} - \sum_{\mathrm{k=1}}^{\mathrm{K}}\mathrm{a_{k,t}^i}, \mathrm{D_{d,t}^i}\right)\mathrm{h_{d,t}^i} - \mathrm{C_F^{d,i}}(\mathrm{t}) - \mathrm{C_1}(\mathrm{t}) - \mathrm{C_2}(\mathrm{t})\right] \tag{14}$$

In local networks, the reward that each retailer earns is

$$r_k(t) = \sum_{i=1}^{M}\left(\min\left(\left(x_{k,t}^i + a_{k,t}^i + e_{k,t}^i\right), D_{k,t}^i\right)h_{k,t}^i - C_F^{k,i}(t) - C_3(t) - C_4(t) - C_5(t) - C_6(t)\right) \tag{15}$$

The objective is to maximize the cumulative total rewards over a finite time horizon $T$, formulated as

$$\max_{\pi_d, \pi_k} \mathbb{E}\left[\sum_{t=0}^{T}\gamma^t\left(r_d(t) + \sum_{k=1}^{K} r_k(t)\right)\right]$$

where $\pi_d$ and $\pi_k$ represents policies for the distribution center and retailers, respectively. The discount factor $\gamma$ determines the importance of future rewards, ensuring long-term optimization rather than optimization of current rewards.

### *4.2 Proposed hybrid A3C-distributed PPO (A3C-DPPO) framework*

Our proposed hybrid A3C-DPPO algorithm builds on the A3C framework (Mnih et al., 2016) and PPO method (Schulman et al., 2017). The A3C component provides quick learning and exploration of the large state–action space, while the PPO component ensures stable policy updates, which is crucial for the sensitive nature of the PSC (Zhang et al., 2019). Here, A3C employs multiple agents running asynchronously to collect experiences from their local environments and compute local gradients for both the policy and value networks based on the collected experiences. Instead of updating the global parameters asynchronously as in A3C, the gradients are sent to a distribution center that functions as a central server as illustrated in Fig. 2. The distribution center synchronizes and aggregates gradients from multiple retailers, ensuring stable and reliable updates using the PPO clipped objective. The core of the hybrid A3C-DPPO algorithm involves updating the policy and value functions for both local and global agents.

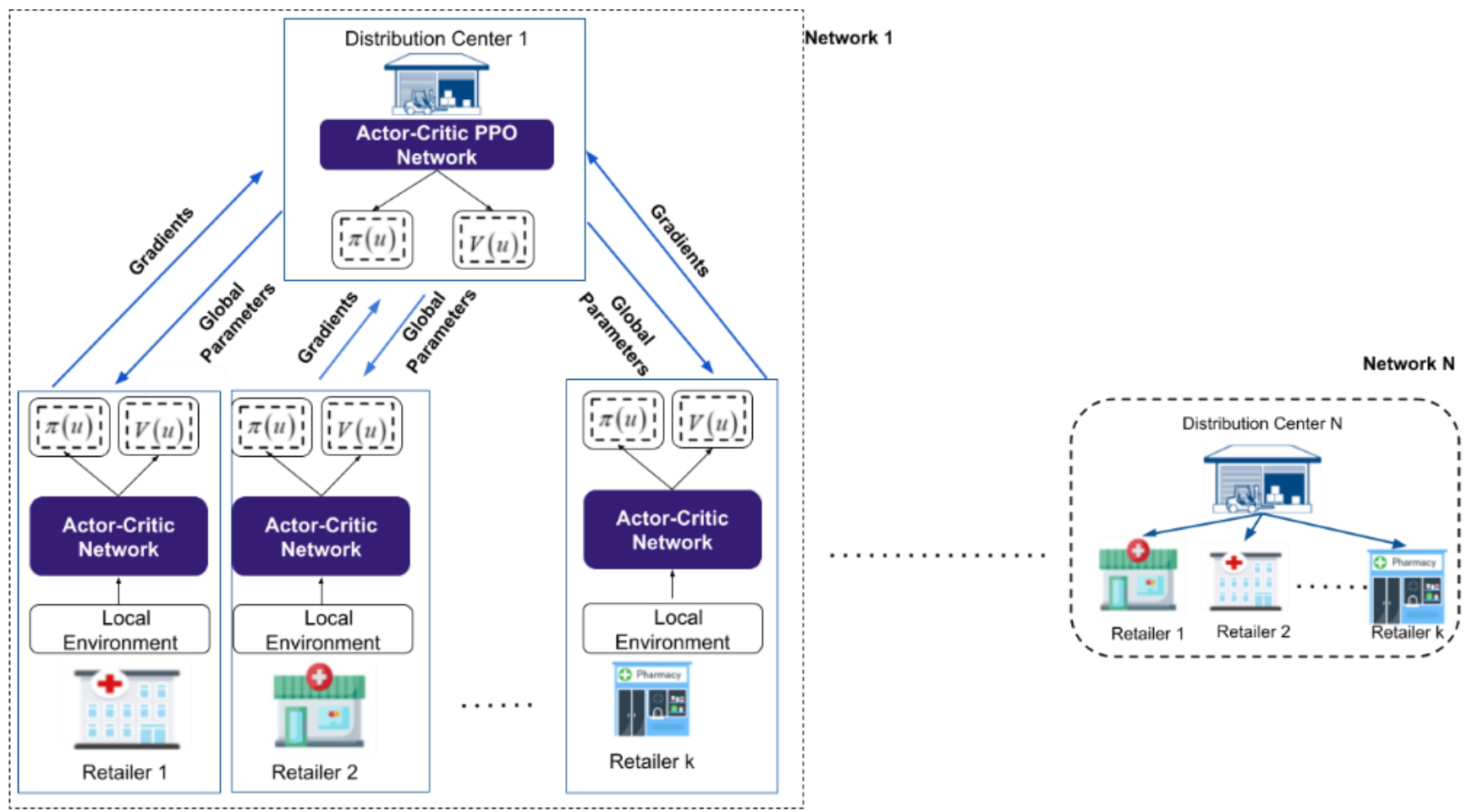

**Fig. 2** Proposed framework of the hybrid A3C-DPPO algorithm

The actor network learns the policy $\pi_{\theta_k}\left(u_k(t)\,|\,v_k(t)\right)$, which determines the probability distribution over actions given the current state $u_k(t)$. The critic network estimates the value function $\vartheta_{\phi_k}\left(u_k(t)\right)$, which evaluates the expected return from a given state $u_k(t)$. The policy update for retailer $k$ aims to maximize the expected reward by adjusting the policy parameters $\theta_k$

$$\theta_k(t+1)=\theta_k(t)+\alpha_a\nabla_{\theta_k}\log\pi_{\theta_k}\left(v_k(t)\,|\,u_k(t)\right)A_k^{\pi} \tag{16}$$

where $\alpha_a$ is the learning rate for the actor network and $\nabla_{\theta_k}\log\pi_{\theta_k}\left(v_k(t)\,|\,u_k(t)\right)$ is the gradient of the log-probability of the action taken. To update the weights of the actor network in the direction that increases the chances of taking better actions, the advantage of taking a certain action over other actions need to be calculated. The advantage function $A_k^{\pi}$ corresponding to each retailer is defined as

$$A_k^{\pi}(t)=\sum_{l=0}^{L}(\gamma\beta)^l\,\delta_{t+l}^k \tag{17}$$

where $\delta_t^k=r_k(t)+\gamma\vartheta_{\phi_k}\left(u_k(t+1)-\vartheta_{\phi_k}\left(u_k(t)\right)\right)$ is the temporal difference error, $\beta\in[0,1]$ is the generalized advantage estimation (GAE) parameter that controls the bias-variance trade-off. The advantage function $A_k^{\pi}(t)$ is computing using a weighted sum of $l$-step temporal difference errors $\delta_t^k$, which balances bias and variance. The critic network update adjusts the value function parameters $\phi_k$ to minimize the difference between the predicted and actual returns as follows

$$\phi_k(t+1)=\phi_k(t)+\alpha_v\left(r_k(t)+\gamma\vartheta_{\phi_k}\left(u_k(t+1)-\vartheta_{\phi_k}\left(u_k(t)\right)\right)\right)\nabla_{\phi_k}\vartheta_{\phi_k}\left(u_k(t)\right) \tag{18}$$

where $\alpha_v$ is the learning rate for the critic network and $\nabla_{\phi_k}\vartheta_{\phi_k}\left(u_k(t)\right)$ is the gradient of the value function with respect to its parameters.

The global agent uses the DPPO approach to update its policies and value functions. The distribution center optimizes overall IM by coordinating actions across all retailers. The actor network for the distribution center determines the policy $\pi_{\theta_d}\left(u(t)\mid v(t)\right)$, which dictates the quantities to order and distribute. The critic network evaluates the value function $\vartheta_{\phi_d}\left(u(t)\right)$, which assesses the expected return from the global state $u(t)$. The policy update for the distribution center ensures stable updates by clipping the policy ratio as

$$\theta_d(t+1)=\theta_d(t)+\alpha_a\mathbb{E}\left[\min\left(\eta_t(\theta_d)\nabla_{\theta_d}\log\pi_{\theta_d}\left(v(t)\mid u(t)\right),clip\left(\eta_t(\theta_d),1-\sigma,1+\sigma\right)A_d^\pi\right)\right] \tag{19}$$

where $A_d^\pi(t)=\sum_{l=0}^{L}(\gamma\beta)^l\,\delta_{t+l}^d$ and $\delta_t^d=r(t)+\gamma\vartheta_{\phi_d}\left(u(t+1)\right)-\vartheta_{\phi_d}\left(u(t)\right)$ represents the advantage function for the distribution center and $\eta_t(\theta_d)$ is the probability ratio between the current and the old policy, calculated as

$$\eta_t(\theta_d)=\frac{\pi_{\theta_d}\left(u(t)\mid v(t)\right)}{\pi_{\theta_d}^{old}\left(u(t)\mid v(t)\right)} \tag{20}$$

When the ratio $\eta_t(\theta_d)>1$, the new strategy is likely to be selected; otherwise, the old strategy continues. This prevents new strategies from significantly deviating from old strategies. Additionally, the adaptive divergence penalty coefficient avoids overestimation of the optimization objective. The term $clip\left(\eta_t(\theta_d),1-\sigma,1+\sigma\right)$ is the clipped probability ratio, which avoids large policy changes. The value function update for the distribution center adjusts parameter $\phi_d$ to minimize the temporal difference error as

$$\phi_d(t+1)=\phi_d(t)+\alpha_v\left(r(t)+\gamma\vartheta_{\phi_d}\left(u(t+1)-\vartheta_{\phi_d}\left(u(t)\right)\right)\right)\nabla_{\phi_d}\vartheta_{\phi_d}\left(u(t)\right) \tag{21}$$

The global actor (distribution center) aggregates gradients from all local actors (retailers). With $k$ local retailers, the cumulative policy gradient for the global actor is calculated as

$$\Delta\theta_d=\frac{1}{K}\sum_{k=1}^{K}o_k\nabla_{\theta_k}\log\pi_{\theta_k}\left(v_k(t)\mid u_k(t)\right)A_k^\pi \tag{22}$$

where $\pi_{\theta_k}$ represents the policy of retailer $k$, parameterized by $\theta_k$; $o_k = \mu_1 e^{-\rho(t-t_k)} + \mu_2 \frac{r_k}{\sum_{k=1}^{K} r_k}$ is an importance weighting factor such that $\mu_1 + \mu_2 = 1$. The term $e^{-\rho(t-t_k)}$ prioritizes recent updates from retailer $k$ and $\frac{r_k}{\sum_{k=1}^{K} r_k}$ ensures that the agent with high average rewards has high importance.

Similarly, the global critic aggregates the gradients from all local agents, and the cumulative value gradient for the global critic is

$$\Delta\phi_d = \frac{1}{K}\sum_{k=1}^{K} o_k \delta_t^k \nabla_{\phi_k} \vartheta_{\phi_k}\left(u_k\left(t\right)\right) \tag{23}$$

where $\vartheta_{\phi_k}\left(u_k\left(t\right)\right)$ represents the value function parameterized by $\phi_k$ for retailer $k$ and $\delta_t^k$ is the temporal difference error, computed as

$$\delta_t^k = r_k\left(t\right) + \gamma\vartheta_{\phi_k}\left(u_k\left(t+1\right) - \vartheta_{\phi_k}\left(u_k\left(t\right)\right)\right) \tag{24}$$

where $r_k\left(t\right)$ is the reward by retailer $k$ at time $t$ and $\vartheta_{\phi_k}\left(u_k\left(t+1\right)\right)$ and $\vartheta_{\phi_k}\left(u_k\left(t\right)\right)$ represent the estimated values of the next and current states, respectively. To ensure stability and synchronization in asynchronous learning process, we incorporate exponential moving averages of global parameters:

$$\theta_d \leftarrow \tau\theta_d + \left(1-\tau\right)\left(\theta_d + \alpha_{a,d}\Delta\theta_d\right) \tag{25}$$

$$\phi_d \leftarrow \tau\phi_d + \left(1-\tau\right)\left(\phi_d + \alpha_{c,d}\Delta\phi_d\right) \tag{26}$$

where $\tau \in \left(0,1\right]$ controls update smoothness and $\alpha_{a,d}$ and $\alpha_{v,d}$ are the learning parameters of the global actor and critic network. After updating the global parameters, each local retailer receives the updated global parameters:

$$\theta_k \leftarrow \theta_d \text{ and } \phi_k \leftarrow \phi_d, \forall k \in \{1,2,.....K\} \tag{27}$$

This ensures that the proposed algorithm effectively manages asynchronous gradient updates while maintaining stability and synchronization in the PSC scenario. The pseudocode summarizing proposed algorithm is outlined in Algorithm 1.

---

**Algorithm 1:** Hybrid A3C-DPPO algorithm for pharmaceutical IM

---

**1: for** each network $n = 1,2,.....,N$ :
**2:** Initialize the parameters $\theta_d$ and $\phi_d$ for the global actor–critic networks
**3:** Initialize the parameters $\theta_k$ and $\phi_k$ for the local actor–critic networks for each retailer
**4:** Initialize the hyperparameters for the actor and critic networks of both local and global agents
**5**: **for** each epoch:
**6:** Reset the cumulative gradients for the global actor $(\Delta\theta_d)$ and global critic $(\Delta\phi_d)$
**7:** **for** each local agent (retailer) $k$ in parallel:

**8**: Simulate the local agent's interaction with the environment to collect state–action–reward–next state tuples
**9:** Compute the advantage function $A_k^{\pi}$ for the local agent
**10:** Calculate the local actor's policy gradient $\nabla_{\theta_k}$
**11:** Compute the temporal difference error $\delta_t^k$ and the local critic's value gradient $\nabla_{\phi_k}$
**12:** Aggregate these gradients into the global cumulative gradients $(\Delta\theta_d)$ and $(\Delta\phi_d)$
**13:** **end for**
**14:** Update the global actor and critic parameters using cumulative gradients.
**15:** Synchronize the updated global parameters with all local agents.
**16:** **end for**
**17**: **end for**

## 5. Numerical experiments

Numerical experiments to evaluate the performance of the proposed A3C-DPPO algorithm for the PSC IM were conducted using a workstation equipped with an Intel® core i9- 12900K CPU, NVIDIA GeForce RTX 4090 GPU 24 GB, and a 64-bit operating system (Windows 11 Professional).

### *5.1 Network structure and problem parameters*

Here, we consider a target PSC network consist of five distribution centers serving ten retailers. The network operates in a stochastic environment with heterogeneous demand arrival rates and replenishment rates across different retailers, and aggregated demand at distribution centers. For presumed statistical distributions, we consider the Normal distribution $D_t^k \sim N(\mu,\sigma^2)$, where $\mu$ and $\sigma$ indicate the mean and the standard deviation of demand. To cover a range of demands, $\left(CV_{nor} = \frac{\sigma}{\mu}\right)$ with different coefficients are considered. For skewed demand scenarios, we consider the Gamma distribution as $D_t^k \sim Gamma(m,\theta)$, with shape parameter $(m)$ and scale parameter $(\theta)$. The mean and variance can be calculated as $\mu = m\theta$ and $\sigma = m\theta^2$, and their relationship is managed using the coefficient of variation $\left(CV_{gamma} = \frac{1}{\sqrt{\mu\sigma}} = \frac{1}{\sqrt{k}}\right)$(Park et al. 2024, Gioia and Minner 2023). Following Peng et al. (2019) and Stranieri et al. (2024), we also map the seasonal demand as $D_t^k = \mu\left(1 + A\times\sin\left(\frac{2\pi t}{T}\right)\right) + \varepsilon(t)$, where $\mu$ is the mean demand, $T$ is the period of seasonality, $A$ is the amplitude of the seasonal fluctuations, and $\varepsilon(t)$ is a random variable representing gaussian noise. Furthermore, to model discrete events (such as medication demand during an epidemic), we consider the Poisson distribution $D_t^k \sim Poisson(\lambda)$ to sample the random demand from the interval $\left[\lambda - z\sqrt{\lambda}, \lambda + z\sqrt{\lambda}\right]$, where $\lambda$ is the average demand rate. The selected demand distributions align with the suggestions of Mohamadi et al. (2024) and Stranieri et al. (2024). Similarly, we randomly sample the varying replenishment lead time with the exponential distribution $\tau_k \sim Exp(\tau)$ for retailers. The range of the replenishment interval is bounded as $[l_{\min}, l_{\max}]$, where $l_{\min}$ and $l_{\max}$ represent the minimum and maximum possible lead times, respectively. The average replenishment rate for distribution center $\tau_d$ is assumed to be 0.1 (Wang and Minner 2024).

Following similar settings (Moor et al. 2022), we consider that all retailers have identical inventory-associated costs, including holding cost, ordering cost, shortage penalty cost and ordering cost, for multiple inventory items in the network. The unit ordering cost and holding cost are fixed at the distribution center and retailer, while the other parameters are varied. We set the emergency cost per item based on the geographical $t_f$ between the distribution center and retailer $k$, i.e., $e^i_{p,k,t} = c_b + t_f . y_{d,k}$, where $c_b$ defines the base penalty cost ($2/item), $t_f$ indicates the transport cost ($1-$2)/item, and $y_{d,k}$ is the distance between the retailer and distribution center.

**Table 3**

Parameters of the PSC

| Parameter | Distribution center | Retailer |
|---|---|---|
| Initial inventory | 100 units/item | 40 units/item |
| Inventory capacity | 1500 units | 300 units |
| Number of inventory items | - | 02 |
| Unit ordering cost | $5/order/item | $10/order/item |
| Unit holding cost | $0.08 per unit/day/item | $0.1per unit/day/item |
| Unit sales price | $3-$7/unit | $5-$10/unit |
| Service penalty cost - For critical medication | $10/unit | $20/unit |
| Service penalty cost - For non- critical medication | $5/unit | $8/unit |

**Table 4**

Test and training sets

| Parameter | Training set | Test set |
|---|---|---|
| Coefficient of variation, $(CV_{nor})$ - Normal distribution | U(0.5, 1.0) | {0.1, 0.3} |
| Coefficient of variation, $(CV_{gamma})$ - Gamma distribution | U(0.45, 1.41) | {0.71, 0.92} |
| Amplitude variation of seasonal distribution $(A)$ | U(0.5, 0.8) | {0.2, 0.9} |
| Noise variation of seasonal distribution $(\varepsilon)$ | U(5, 8) | {2, 9} |
| Poisson demand arrival rate $(\lambda)$ | U(0.5, 20) | {0.3, 5} |
| Replenishment rate $(\tau_k)$ | U(0.05, 0.5) | {0.01, 0.8} |

We assume a maximum distance of 20 km for each distribution center–retailer pair. The unit disposal cost for the distribution center and retailers is set as ($2-$5)/item based on shipment cost and geographic location of the disposal area. Additionally, service penalty costs differ between critical and non-critical medications, reflecting the importance of ensuring availability. The details of these problem parameters are summarized in Table 3.

Using the above settings, we generate data and split it into test and training sets to evaluate the performance of the proposed algorithm. Table 4 summarizes the parameters used for the test and training sets. We consider a range of values to generate the maximum number of combinations, each corresponding to a unique problem instance. The coefficients of variation for Normal and Gamma distributions are chosen based on established demand patterns, where values closer to 1 represent high variability in pharmaceutical demand (Graves & Willems, 2003). The amplitude and noise variations in the seasonal distribution are selected to simulate fluctuating demand cycles commonly observed for seasonal medications (Simchi-Levi et al., 2014). The range of variation of the Poisson demand arrival rate captures both frequent and infrequent demand conditions. The replenishment rates are derived from practical lead-time variations observed in hospital pharmacies, where stock replenishment intervals can range from a few days to a week depending on inventory turnover rates. The hyperparameters are selected through an extensive grid search process to ensure optimal performance. The discount factor is initially explored within the range of 0.49 to 0.99 and fine-tuned to 0.99 for optimal performance, aligning with the recommendations of Gijsbrechts et al. (2022). The hyperparameters of the proposed algorithm are presented in Table 5.

**Table 5**

Hyperparameters of the hybrid A3C-DPPO algorithm

| **Hyperparameter** | **Global agent** | **Local agents** |
|---|---|---|
| Hidden layers | 2 | 2 |
| Neurons in 1st hidden layer | 256 | 128 |
| Neurons in 2nd hidden layer | 128 | 64 |
| Learning rate of actor | 0.00005 | 0.0001 |
| Learning rate for critic | 0.0001 | 0.00001 |
| Discount factor $(\gamma)$ | 0.99 | |
| Clipping parameter $(\sigma)$ | 0.2 | |
| GAE parameter $(\beta)$ | 0.95 | |
| Batch size | 64, 128 | |
| Optimizer | Adam | |
| Episodes | 5000 | |

## 5.2 Numerical results

### 5.2.1 Convergence analysis

We first assess the impact of demand distributions and PSL constraints on the convergence performance of the proposed hybrid A3C-DPPO algorithm. To investigate the long-term effects of the proposed replenishment policy, we run simulations for 5000 episodes to evaluate the model's adaptability and stability under varying conditions. Each episode consists of a sequence of time steps taken to reach a terminal state, where the agent interacts with the environment to learn the optimal policy through trial and error. We initially tune the parameters of the proposed algorithm with the Gamma distribution and then compare its performance with those obtained with the Normal and Poisson distributions under varying demand scenarios. As illustrated in Fig. 3(a), with the Normal distribution, the algorithm converges quickly, with higher rewards, due to the stable demand pattern, which simplifies learning and accelerates convergence. By contrast, the Gamma distribution introduces demand variability, and the additional learning required to adapt to fluctuating demand conditions results in slower convergence and moderate rewards. The unpredictable behavior in the presence of the Poisson distribution leads to slower convergence and lower rewards due to the increased difficulty of inventory optimization. The results demonstrate that the proposed algorithm can effectively adapt to various demand patterns, although stable demand patterns, such as those seen in the Normal distribution, yield faster convergence and improved performance. Additionally, Fig. 3(b) illustrates the performance of the proposed algorithm under varying minimum target fill rates $\left(\beta_{k,\min}^{i}\right)$. As the fill rate increases, the algorithm needs to maintain higher inventory levels to meet the stricter service requirements, which limits its ability to optimize costs and rewards effectively. This increased constraint on IM leads to slower convergence and lower average rewards because the algorithm has less flexibility to adjust inventory based on demand fluctuations. By contrast, lower fill rate requirements allow the algorithm to optimize inventory more precisely, resulting in faster convergence and higher rewards.

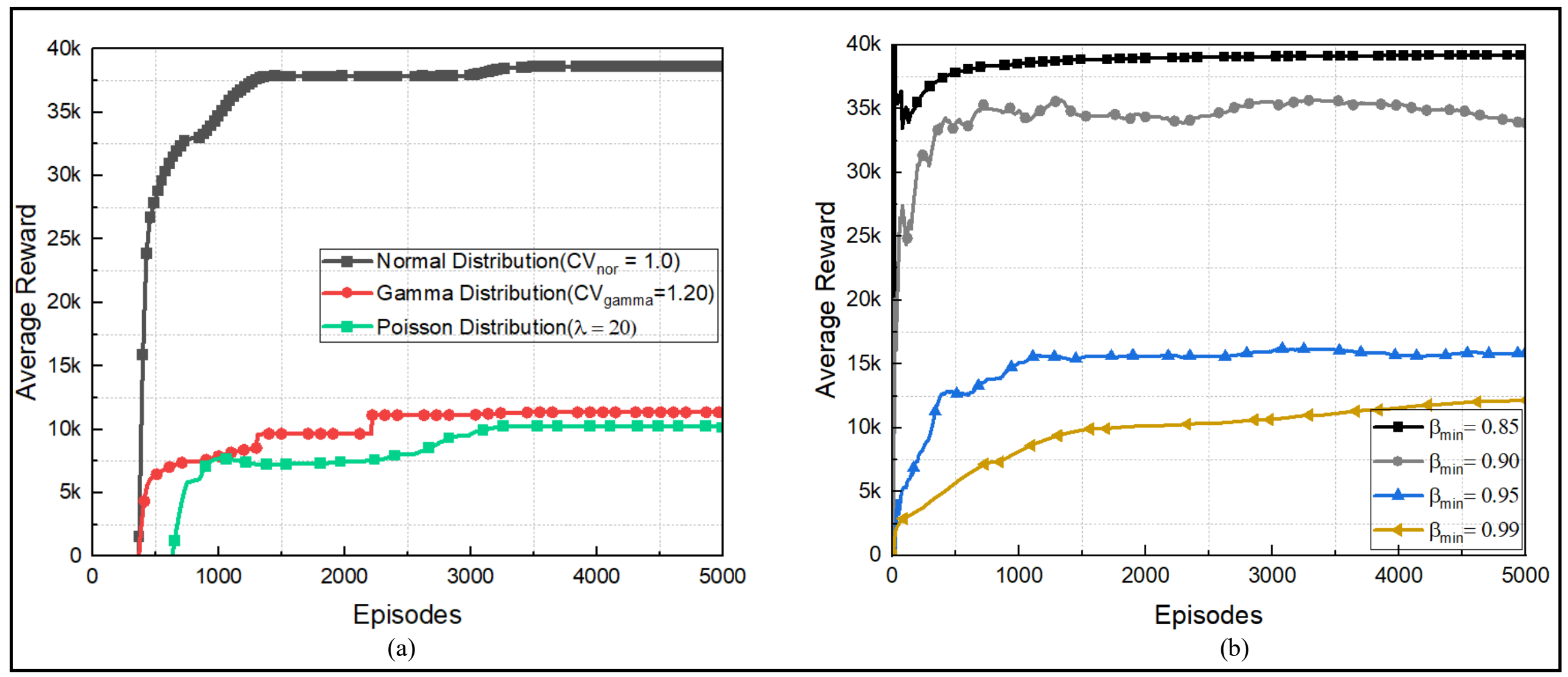

**Fig. 3** Convergence analysis with different (a) demand distributions (b) target fill rates

### *5.2.2 Comparison with baseline policies*

In this section, we compare the performance of the hybrid A3C-DPPO algorithm with that of the baseline policies in the PSC environment. As shown in Fig. 4(a), the proposed algorithm results in higher average rewards than the baseline policies under different demand distributions. During this experiment, the replenishment rate and minimum target fill rate are fixed at $\tau_k = 0.5$ and $\beta_{i,\min} = 0.90$, respectively, for all pharmaceutical products available in the inventory. To make ordering decisions, the classical $(s,S)$ inventory replenishment policy relies only on the current position of inventory items, potentially undermining responsiveness under uncertain demand. DQN, which incorporates state space information for improved decision-making, performs better than the $(s,S)$ policy, but its reliance on a single Q-network limits its ability to handle complex non-linear demand patterns, resulting in unstable convergence under Gamma and Poisson distributions (Fig. 4(b)).

PPO demonstrates improved learning stability through optimized policy gradients, but its sample inefficiency poses challenges, especially in environments with complex demand patterns.

The hybrid A3C-DPPO algorithm overcomes these limitations by employing distributed learning with dynamic policy updates across both local (retailer) and global (distribution center) agents. The asynchronous actor–critic architecture enables the algorithm to learn faster and optimize effectively across diverse demand scenarios. As shown in Fig. 4(b), the hybrid A3C-DPPO algorithm achieves optimal rewards in approximately 3000 episodes, while PPO and DQN require over 4500 episodes to achieve comparable performance. Furthermore, under the Gamma distribution ( $CV_{gamma}$ = 0.92), the average rewards of the hybrid A3C-DPPO algorithm are 28% and 35% higher than those of PPO and DQN, respectively. The hybrid algorithm also demonstrates improved stability in handling complex demand patterns, achieving convergence with minimal variance.

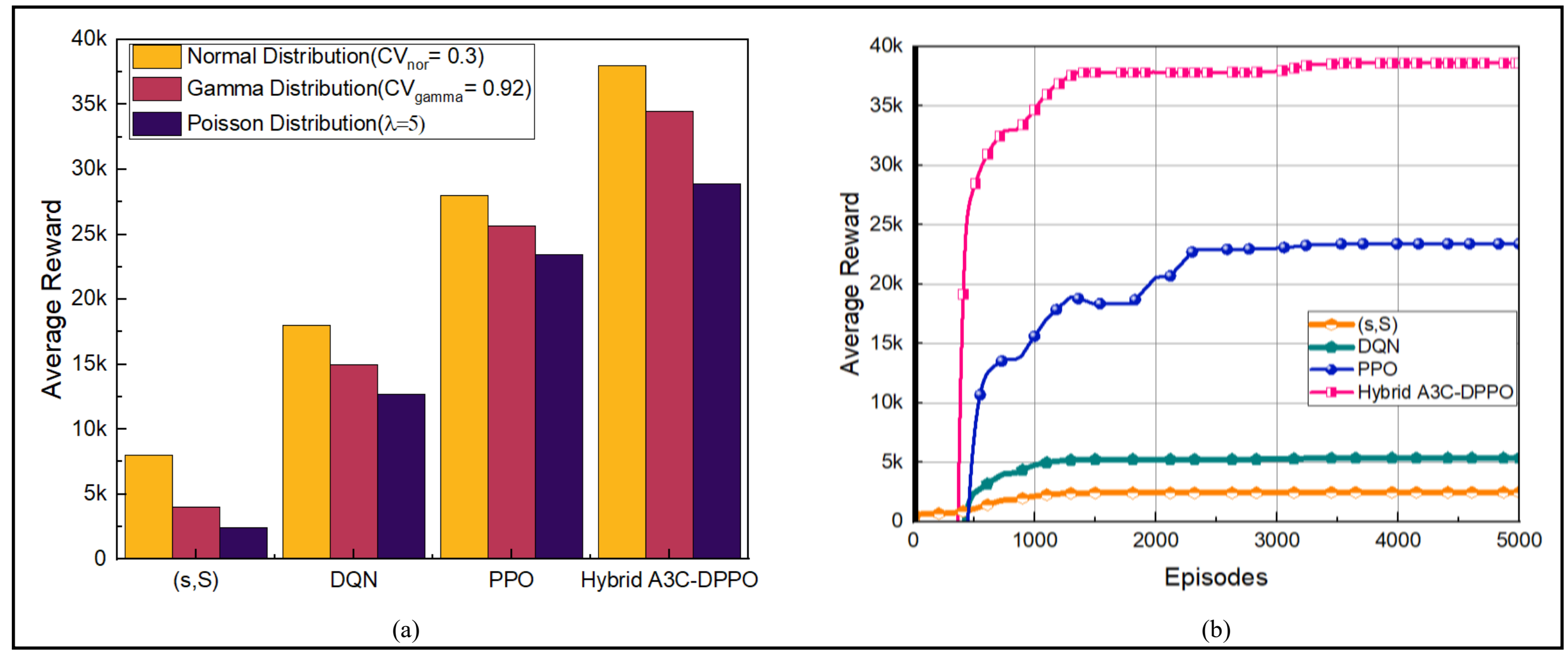

**Fig. 4** Performance comparison in terms of (a) average reward (b) convergence analysis of the proposed A3C-DPPO with baselines algorithms

Table 6 reports the performance of the hybrid A3C-DPPO algorithm in terms of average reward under different demand distributions, target fill rates, and replenishment rates. As the target fill rate increases, the average reward generally decreases across all distributions due to stricter PSL constraints. These constraints necessitate the maintenance of higher inventory levels, which reduces the flexibility of the proposed algorithm in optimizing rewards. The impact of demand variability is evident across different distributions. The Gamma and Poisson distributions, which involve higher variability in $CV_{gamma}$ and $\lambda$, result in pronounced declines in the average reward compared to the Normal distribution. By contrast, seasonal demand distributions require increased inventory levels to accommodate peaks in demand, leading to higher inventory costs and lower overall rewards. Additionally, replenishment rates significantly influence rewards. Higher replenishment rates facilitate more frequent restocking, helping buffer against demand variability and leading to improved average rewards.

**Table 6**

Analysis of the performance of the hybrid A3C-DPPO algorithm in terms of average reward under different test instances

| Target fill rate ($\beta^i_{k,\min}$) | Metric (Average Reward) | | | | | | | | | | | |
|---|---|---|---|---|---|---|---|---|---|---|---|---|
| | Normal distribution $\left(CV_{nor}=\frac{\sigma}{\mu}\right)$ | | Gamma distribution $\left(CV_{gamma}=\frac{1}{\sqrt{\mu\sigma}}=\frac{1}{\sqrt{k}}\right)$ | | Seasonal distribution $\mu\left(1+A\times\sin\left(\frac{2\pi t}{T}\right)\right)+\varepsilon(t)$ | | | | Poisson distribution ($\lambda$) | | Replenishment rate ($\tau_k$) | |
| | $CV_{nor}=$ 0.1 | $CV_{nor}=$ 0.3 | $CV_{gamma}=$ 0.71 | $CV_{gamma}=$ 0.92 | $A=0.2$ | $A=0.9$ | $\varepsilon=2$ | $\varepsilon=9$ | $\lambda=0.3$ | $\lambda=5$ | $\tau=0.01$ | $\tau=0.8$ |
| 0.85 | 37,560 | 37,260 | 36,780 | 34,600 | 35,600 | 27,600 | 29,700 | 24,800 | 33,500 | 29,700 | 32,700 | 36,700 |
| 0.90 | 34,800 | 33,800 | 33,400 | 32,400 | 32,800 | 25,900 | 24,700 | 22,700 | 29,900 | 26,900 | 25,000 | 33,500 |
| 0.95 | 31,500 | 30,300 | 29,800 | 27,800 | 28,600 | 20,300 | 21,200 | 19,300 | 26,300 | 23,100 | 22,300 | 29,800 |
| 0.99 | 29,700 | 28,000 | 24,112 | 22,800 | 23,980 | 18,600 | 19,500 | 16,800 | 24,300 | 19,200 | 17,000 | 24,100 |

**Table 7**

Comparison of the performance of the hybrid A3C-DPPO algorithm with baseline policies under different test instances in terms of average reward

| $\beta^i_{k,\min}$ | Algorithm | $CV_{nor}=0.3$ | $CV_{gamma}=0.71$ | $A=0.9$, $\varepsilon=9$ | $\lambda=0.5$ |
|---|---|---|---|---|---|
| 0.85 | $(s,S)$ | 10,500 | 9,780 | 5,500 | 6,900 |
| | DQN | 14,100 | 12,780 | 14,700 | 19,600 |
| | PPO | 28,200 | 26,900 | 18,400 | 22,560 |
| | Hybrid A3C-DPPO | 37,260 | 34,600 | 24,610 | 32,800 |
| 0.90 | $(s,S)$ | 10,520 | 9780 | 5,500 | 6,900 |
| | DQN | 20,600 | 13,900 | 12,700 | 17,300 |
| | PPO | 26,900 | 23,100 | 15,000 | 21,893 |
| | Hybrid A3C-DPPO | 33,900 | 32,400 | 22,000 | 30,900 |
| 0.95 | $(s,S)$ | 7,800 | 5,890 | 3,800 | 5,100 |
| | DQN | 17,300 | 16,600 | 10,500 | 14,300 |
| | PPO | 23,400 | 20,800 | 14,700 | 18,600 |
| | Hybrid A3C-DPPO | 30,200 | 28,800 | 19,833 | 25,600 |
| 0.99 | $(s,S)$ | 5,200 | 4,200 | 2,700 | 3,800 |
| | DQN | 12,200 | 13,500 | 8,300 | 12,800 |
| | PPO | 19,900 | 17,400 | 18,800 | 16,782 |
| | Hybrid A3C-DPPO | 22,800 | 22,800 | 16,890 | 20,893 |

Table 7 presents a comprehensive comparative performance analysis of the hybrid A3C-DPPO algorithm against baseline policies across different demand distributions and target fill rates. As shown in Table 7, the hybrid A3C-DPPO algorithm consistently outperforms the other algorithms across all target fill rates and demand distributions, demonstrating its robustness and flexibility in handling various demand patterns. The optimality gap values indicate that the hybrid A3C-DPPO algorithm achieves better performance with lower deviation from the optimal solution, highlighting its adaptability in managing inventory under dynamic conditions.

Additionally, we investigate the impact of various inventory cost components on the performance of the proposed algorithm across varying PSLs. As illustrated in Fig. 5, as the PSL increases from 0.85 to 0.99, the average inventory cost also increases. Higher PSLs require more inventory to reliably meet demand, increasing the holding cost. Compared with the holding cost, the replenishment cost increases less steeply with increasing PSL, indicating that the hybrid A3C-DPPO algorithm may optimize replenishment decisions more efficiently, with replenishment occurring just in time to balance PSLs and costs. Greater PSLs (approaching 0.99) result in a high stockout penalty, i.e., the penalty for unmet demand increases. These results indicate that maintaining a higher PSL requires substantial investment to avoid penalties related to non-availability.

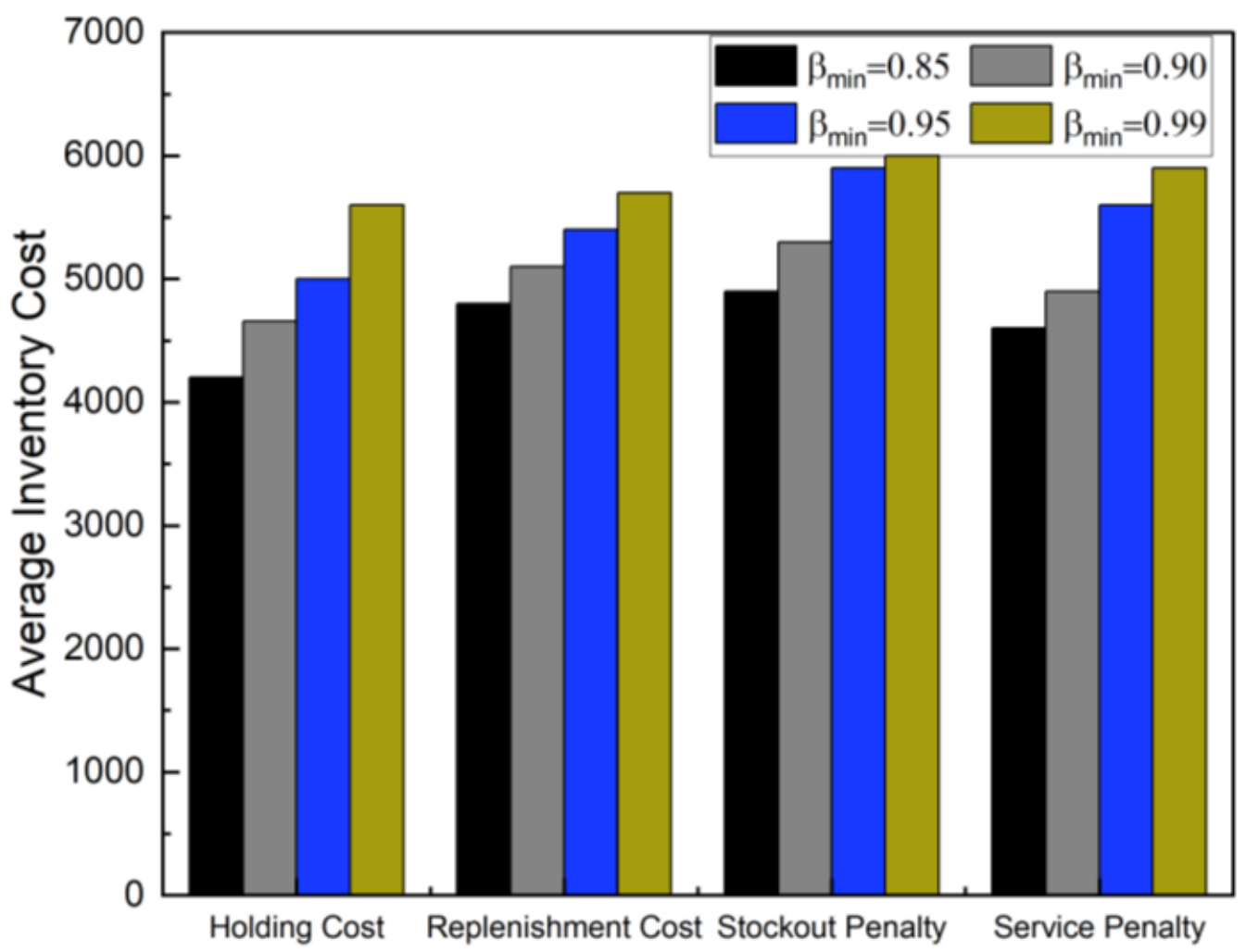

**Fig. 5** Sensitivity analysis of the impact of cost components on algorithm performance

## *5.3 Numerical validation with a real-world pharmaceutical inventory system*

We conduct a numerical validation of the proposed algorithm using data from a hospital information system that acts as a distribution center for two pharmacies within the hospital. The data are collected over a period of 52 weeks and include date, time, pharmacy location, medication name and medication quantity. The collected data represent the details of 2447 unique medications. We collect information on medications belonging to specific categories, namely, anti-diabetic, anti-viral and anti-pyretic medications ordered by the pharmacies. These categories are selected to assess the performance of the proposed algorithm under different demand distributions: seasonal, non-seasonal, and pandemic. The best-fitted demand distributions align with the suggestions of research and the medical literature (Saha & Rathore, 2024; Mohamadi et al., 2024). The names of the pharmacies and hospital are not disclosed due to confidentiality. The ordering cost, holding cost, shortage penalty, disposal cost and inventory level of each medication at the pharmacies and distribution center were determined based on discussions with the pharmacies and are presented in Table 8. Since these medications are crucial for patient treatment, pharmacies should ensure their availability; consequently, a minimum target fill rate of 0.99 is set for each medication at both pharmacies.

### 5.3.1 Performance comparison

The aim of the numerical experiments using real-world data is to confirm the suitability of the proposed algorithm for inventory replenishment under varying demand scenarios. For comparison, we consider the baseline RL algorithms DQN and PPO. Table 9 presents the performance comparison in terms of profitability as average reward and average inventory cost. The proposed inventory replenishment policy indicates that the A3C-DPPO algorithm learns in response to dynamic demand while significantly reducing inventory costs. As shown in Table 9, the average inventory cost is higher for Tamiflu 30mg TAB and Spikevax 2.5mL INJ than for Metformin 500mg TAB. However, the proposed algorithm achieves substantial cost savings across all medications, reducing inventory costs by 61.5% and 37.3% for Tamiflu 30mg TAB compared to DQN and PPO, respectively. Similarly, for Metformin 500mg TAB, the A3C-DPPO algorithm reduces the cost by 95.1% compared to DQN and 82.8% compared to PPO, while for Spikevax 2.5mL INJ, the reductions are 65.7% and 29.5%, respectively. In terms of average reward, the proposed A3C-DPPO approach consistently outperforms the baseline DQN and PPO methods, improving rewards by 23.1% and 13.5% for Tamiflu 30mg TAB and 48.2% and 28.3% for Metformin 500mg TAB, respectively. All three algorithms are MDP based, but the proposed algorithm shows superior performance under varying demand scenarios.

**Table 8**

Inventory parameters of pharmaceutical network

| Medication | Category (Demand Distribution) | Inventory Level | $c_o$ (/unit) | $c_h$ (/unit/week) | Penalty $(w_i)$ | $c_p$ (/unit) | $c_\varsigma$ (/unit) |
|---|---|---|---|---|---|---|---|
| **Pharmacy 1 (retailer)** | | | | | | | |
| Tamiflu 30mg TAB | Type I: (Seasonal variation) | 50 | \$5 | \$0.5 | \$10 | \$6 | \$0.01 |
| Metformin 500mg TAB | Type II: (Non-seasonal) | 20 | \$7 | \$1 | \$8 | \$8 | \$0.2 |
| Spikevax 2.5mL INJ | Type III: (Pandemic) | 25 | \$9 | \$0.75 | \$12 | \$10 | \$0.05 |
| **Pharmacy 2 (retailer)** | | | | | | | |
| Tamiflu 30mg TAB | Type I: (Seasonal variation) | 40 | \$3 | \$2 | \$12 | \$8 | \$0.02 |
| Metformin 500mg TAB | Type II: (Non-seasonal) | 35 | \$6 | \$0.5 | \$10 | \$9 | \$0.3 |
| Spikevax 2.5mL INJ | Type III: (Pandemic) | 20 | \$9 | \$1 | \$15 | \$12 | \$0.1 |
| **Distribution Center** | | | | | | | |
| Tamiflu 30mg TAB | Type I: (Seasonal variation) | 1000 | \$1 | \$0.25 | \$4 | - | - |
| Metformin 500mg TAB | Type II: (Non-seasonal) | 800 | \$2 | \$0.08 | \$5 | - | - |
| Spikevax 2.5mL INJ | Type III: (Pandemic) | 750 | \$8 | \$0.25 | \$7 | - | - |

**Table 9**

Performance comparison with the real-world dataset

| Medication | Average Reward | | | Average Cost | | |
|---|---|---|---|---|---|---|
| | **DQN** | **PPO** | **Proposed** | **DQN** | **PPO** | **Proposed** |
| Tamiflu 30mg TAB | \$42767.22 | \$46431.76 | \$52660.48 | \$5986 | \$3675 | \$2306 |
| Metformin 500mg TAB | \$33567.99 | \$38772.00 | \$49742.00 | \$4524 | \$1278 | \$220 |
| Spikevax 2.5mL INJ | \$11787.20 | \$22312.63 | \$30399.33 | \$9900 | \$4375 | \$3085 |

DQN suffers from overestimation and instability in high-dimensional state spaces, leading to sub-optimal decision-making in dynamic environments. PPO optimizes policy updates through a clipped objective function at the cost of limited exploration, hindering adaptability in highly dynamic settings. By contrast, the inventory policy with the proposed algorithm utilizes parallel actor learners to explore the environment in parallel, increasing training stability and improving policy robustness compared to the single-threaded DQN model. Our results align with those of Mnih et al. (2016) and Schulman et al. (2017).

### *5.4.2 Sensitivity analysis of demand parameters*

Additionally, we conduct a sensitivity analysis to investigate the performance of the proposed algorithm under varying demand conditions. The demand variance represents the variability or spread in demand over time and indicates how much the current demand deviates from the average demand. Mathematically, the demand variance can be calculated as

$$\mathrm{var}(D) = \frac{1}{J}\sum_{j=1}^{J}\left(D_j - \mu_D\right) \tag{28}$$

where $J$ is the total number of observations (52 weeks). Fig. 6 illustrate the performance of the proposed algorithm as a function of the demand variance. Unlike other inventory policies, the proposed A3C-DPPO policy is highly robust against demand fluctuations, which is advantageous for IM systems that require timely decision-making. By integrating both actor–critic and distributed parallel learning, A3C-DPPO can adapt to varying levels of demand volatility while maintaining a balance between exploration and exploitation. This allows it to optimize replenishment decisions more effectively, leading to higher rewards and lower inventory costs, even under increasing demand variability. By contrast, traditional methods, such as $(s,S)$, struggle with demand uncertainty, and DQN and PPO are less efficient in adapting to high-variance environments due to their slower convergence and limited capacity for decentralized decision-making.

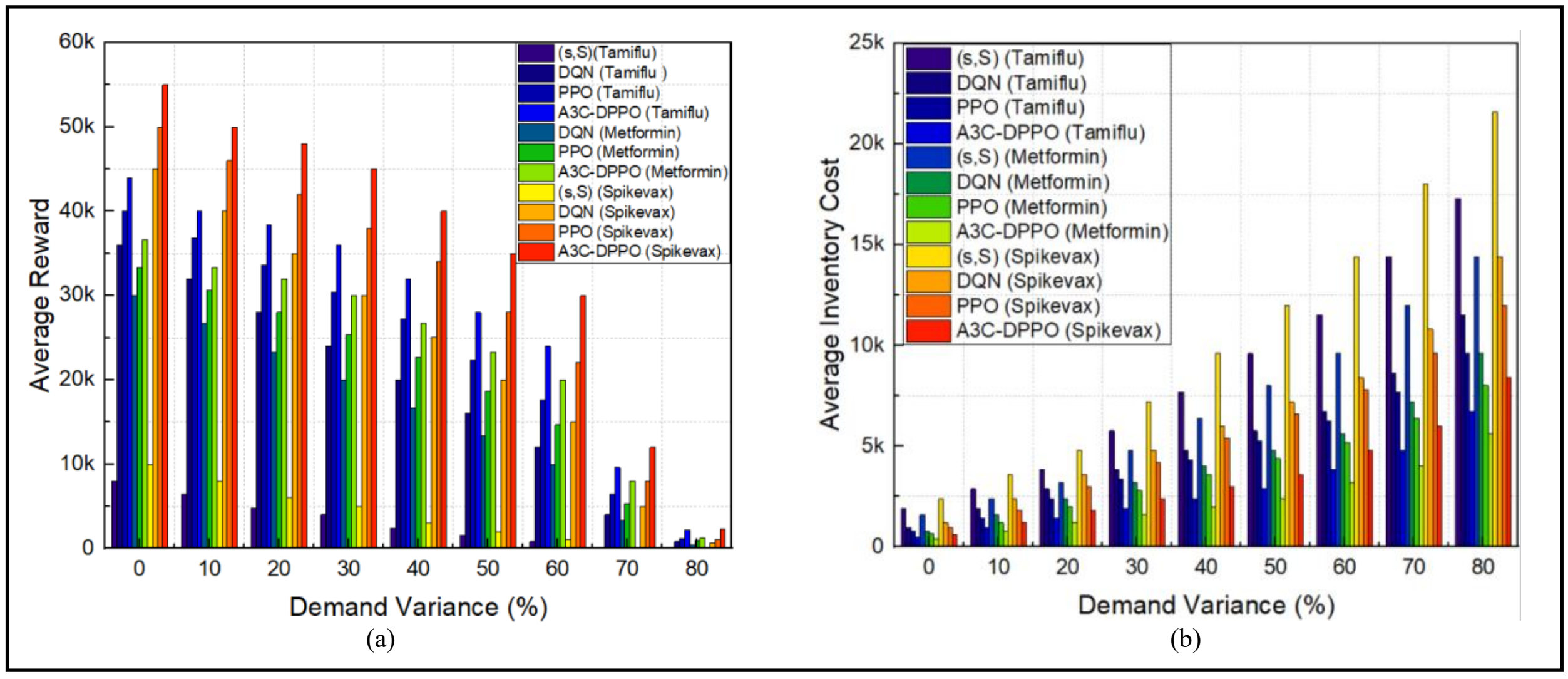

**Fig. 6** Effect of demand variance on (a) average reward (b) average inventory cost

## 6. Conclusion

This research advances PSCs by proposing a dynamic inventory optimization framework that quickly responds to fluctuating demand through intelligent decision-making and delivers high PSLs. Our approach employs A3C-DPPO, a hybrid DRL algorithm that combines value-based and policy-based approaches, to optimize IM along the PSC. This hybrid method can handle the large state–action space created by the dynamic, volatile PSC context. The framework successfully adapts to fluctuating demand, mitigates shortages, eliminates excess stock and reduces costs. We validate the performance of the proposed algorithm in terms of total inventory cost, including ordering cost, holding cost, stockout penalty, and disposal cost, under PSL constraints. Moreover, its performance across different demand scenarios, including seasonal and non-seasonal variations, is benchmarked against contemporary policies. The numerical results and validation with real-world scenarios give rise to the following key findings:

- Maintaining a high PSL and responsiveness in the PSC is crucial, particularly for medications with highly uncertain demand. Our proposed hybrid DRL algorithm optimally selects the required number of medications, ensuring effective IM within a complex PSC. Its performance is validated in numerical experiments conducted under various demand distributions and target fill rates. Notably, the A3C-DPPO algorithm performs significantly better than traditional methods under high demand variability. Furthermore, sensitivity analysis across different target fill rates shows that the proposed algorithm consistently balances PSL and cost efficiency, ensuring effective IM.
- The novel integration of A3C and DPPO harnesses the complementary strengths of the two methods to enhance learning stability and efficiency under uncertain demand conditions. This hybrid architecture not only accelerates convergence but also enables seamless adaptation to a wide range of demand distributions, marking a significant step forward in robust, real-time decision-making for PSCs.

Future researchers can extend the model to consider fully connected distribution centers and retailers. Other potential lines of investigation include temperature-sensitive pharmaceutical products with specific sets of requirements along the PSC and the evolution of cooperative interactions among retailers (e.g., hospitals and pharmacies) to share pharmaceutical products with the aim of reducing shipment costs and enhancing responsiveness.